\documentclass{article}
\usepackage{IGARSS_style,amsmath,epsfig}
\usepackage{amssymb}
\usepackage{multirow}
\usepackage{xcolor}
\usepackage{graphicx}
\usepackage{enumitem} 
\usepackage[caption=false]{subfig} 
\usepackage[export]{adjustbox} 
\usepackage{hyperref}
\hypersetup{colorlinks=true,linkcolor=blue,filecolor=magenta,urlcolor=blue,}
\usepackage{amssymb}
\usepackage{mathtools}

\title{Channel-based attention for LCC using Sentinel-2 time series}

\newcommand\blfootnote[1]{%
  \begingroup
  \renewcommand\thefootnote{}\footnote{#1}%
  \addtocounter{footnote}{-1}%
  \endgroup
}

\twoauthors
  {H. Courteille, A. Benoit, N. Méger, A. Atto}
	{Polytech Annecy-Chambery, LISTIC \\
	Université Savoie Mont Blanc\\
	 F-74940 Annecy, France }
  {D. Ienco }
	{INRAE,	UMR TETIS\\
	Université de Montpellier,\\
	F-34000 Montpellier, France}

\begin{document}


\maketitle
\begin{abstract} 

\blfootnote{This work was supported by CNES. It is based on
observations with the MultiSpectral Instrument embarked on Sentinel-2 and 
funded by CNES-TOSCA project \textit{START Deep} under grant 6177/5930.}
Deep Neural Networks (DNNs) are getting increasing attention to deal with Land Cover Classification (LCC) relying on Satellite Image Time Series (SITS). Though high performances can be achieved, the rationale of a prediction yielded by a DNN often remains unclear. An architecture expressing predictions with respect to input channels is thus proposed in this paper. It relies on convolutional layers and an attention mechanism weighting the importance of each channel in the final classification decision.
The correlation between channels is taken into account 
to set up shared kernels and lower model complexity.
Experiments based on 
a Sentinel-2 SITS
show promising results. 


\end{abstract}
\begin{keywords}
Deep Learning, Attention, Land Cover Classification, Satellite Image Times Series, Multivariate Time Series, Sentinel-2
\end{keywords}

\section{Introduction}\label{sec:intro}


DNNs are identified as key methods for data-driven Earth system science~\cite{Reichstein2019DeepLA}. They are indeed able to extract complex spatiotemporal features from geospatial data streams such as SITS, which are nowadays widely available thanks to the development of Earth observation programmes such as Landsat~\cite{WULDER2019127} or Copernicus~\cite{Copernicus}. For example, 
the Copernicus Sentinel-2 mission 
freely delivers optical images of any location every 5 days on average.
Such SITS, when processed with DNNs, empower remote sensing applications such as climate surveillance, agriculture monitoring or LCC~\cite{WULDER2019127,Pelletier,IencoRec}. \\
However, while reaching high performances, the rationale leading to predictions is basically not made available by DNNs, that, in turn, are often considered as \textit{black box} methods~\cite{Guidotti2018}. Original methods have thus been designed to \textit{open} such black boxes and explain their predictions. An interesting survey and classification of these methods can be found in~\cite{Guidotti2018}. Some of these methods focus on explaining each one of the outcomes, which can be done by providing the subset of the data that is mainly responsible for the prediction, i.e. a \textit{Saliency Mask} (SM). Identifying such a SM can be performed by DNNs themselves using an \textit{attention} mechanism. A good example can be found in~\cite{attention}, where a DNN automatically generates image captions from words associated to detected salient visual objects.
This mechanism has also been proven to be useful for LCC when using SITS. For instance, in~\cite{IENCOTassel}, the spatial components responsible for a prediction are made available. As shown in~\cite{Garnot} and~\cite{RUWURM2020421}, the temporal components leading to a prediction can also be successfully identified and provided as a rationale.\\ 
In this paper, a DNN architecture allowing to directly interpret its own predictions according to input channels is proposed. It relies on an attention-based mechanism that is fed by features obtained from each channel through temporal convolution-based models. Moreover, the correlations between the different channels are taken into account by using shared kernels to lower model complexity and increase the overall accuracy. The proposed architecture has been selected from a pool of potential neural structures tested on a Sentinel-2 SITS LCC task in terms of classification performances and model complexity levels.
Experiments show that the retained architecture achieves good performances when compared to state-of-the-art approaches and that channel-based explanations are meaningful. 

\section{Architectural settings}

\subsection{Guidelines}
As reported in recent works such as \cite{Pelletier, IENCOTassel}, the expressivity of deep Convolution Neural Networks (CNNs) enables great performances for SITS LCC tasks relying on fully supervised model learning. Other approaches based on recurrent cell models such as Long Short Term Memory (LSTM)
cells (e.g.,~\cite{IencoRec}) could also be considered. However, the potentially large temporal extent of the filters has limited interest for short time series while being more difficult to explain and train. In the following, an overview of CNN-based recent contributions is provided to position our contribution.

The \textit{TASSEL}~\cite{IENCOTassel} model works at the object level by 1) segmenting a reference image of the SITS and identifying the spatiotemporal clusters contained within each segment, 
and 2) training a CNN on the temporal dimension to extract relevant features and then classify each object. Each prediction is simultaneously explained by an attention mechanism weighting the importance of each centroid in the final decision. 
Such an explanation is visualized by assigning obtained weights to the spatial footprint of the clusters defined by the centroids. 
Though this approach allows to reduce the number of parameters of 
the CNN while providing users with handy spatial explanations, a pixel-based model is here preferred to avoid any spatial preprocessing and propose an end-to-end model.\\
Such a strategy is for example adopted by the \textit{TempCNN} model~\cite{Pelletier} where different convolutional blocks 
are applied on both the temporal and the spectral domains.
Interestingly, though spatial information is ignored, high classification performances are obtained. \textit{TempCNN} predictions are nevertheless not explained.

Inspired by \textit{TASSEL} and \textit{TempCNN}, different end-to-end pixel-based architectures incorporating an attention mechanism are proposed and evaluated in this paper. Whatever the proposed architecture, the attention mechanism is designed to explain predictions with respect to input channels. Explanations relying on input channels have been shown to be meaningful in~\cite{campos} where they are obtained using an added noise permutation approach. It is here proposed to rely on an attention mechanism to avoid making any assumption about noise permutation features.
To our knowledge, though relevant for LCC when using a single image (e.g., \cite{rs11020159}), channel-based attention has never been adopted for STIS LCC. 
Different architecture settings are here considered to check  whether convolutions should be factored for correlated channels or not, and  whether attention should be applied before classification or along an auxiliary task branch. All architectures follow the same workflow by first extracting features with convolutions and then classifying with respect to an attention module.

\subsection{Convolutional blocks} 
As regards spectral correlations, three types of features extraction are proposed and listed hereafter. Each approach relies on convolutions applied onto the temporal dimension and each channel is processed separately.\\ 
\textbf{(A)} \label{it:archIndepA}  \textbf{channel specific convolution blocks:}  kernels are dedicated to each channel such that a specific representation is identified at the cost of increased model complexity.\\
\textbf{(B}) \label{it:archShareB} \textbf{shared convolutions:} correlated channels are grouped and processed by the same kernels. Extracted features of the same group thus rely on the same process. As a side effect, model complexity is reduced.\\
\textbf{(C)} \label{it:archiMultiC} \textbf{multistep features extraction:} shared convolutions are first applied on channel groups. Resulting features are then processed in a unified way by a common set of layers.\\
For all the proposed models, 2D convolutions with kernel size $(k, 1)$ are considered to permit shared 1D convolution in an efficient way, taking advantage of computation parallelism. 
Further, two convolution cascade strategies are investigated:\\
    \textbf{(i)} \textbf{a multi-scale  approach} with sub-sampling and an increase of the number of features. This classical approach  yields a high number of kernels and a high number of parameters.\\
    \textbf{(ii)}  \textbf{a processing at the original scale}   with no sub-sampling and a decrease in the number of features. As proposed in~\cite{Pelletier}, this approach significantly limits the number of parameters and allows to keep information resolution all along the process. Kernel size can be extended to deal with large fields of view. High performance levels can be obtained with this strategy for SITS LCC~\cite{Pelletier}.\\
\subsection{Channel attention} \label{subsec:ChannelAtt} For each input sample, related features outing from the convolution block are of shape $(N_{feat}, B)$ with B, the initial number of bands and $N_{feat}$ the number of features arranged as a flat vector. 
Let $h_1,..,h_B$ denote the $N_{feat}$-dimensional feature vectors obtained from each one of the $B$ channels. According to the additive attention detailled in \cite{Bahdanau2015},  the channel weight $\alpha_i$ is computed as follows:
$$ \alpha_i = sigmoid\left( < u ,  tanh(W h_i +b ) >\right) \text{ for } 1\le i \le B $$
where $W$ and $b$ respectively denote the weights and bias of a dense layer, and $u$ is a vector of parameters. All these parameters are learnt. Instead of using softmax as in ~\cite{Bahdanau2015}, weight summation assumption is relaxed by employing a sigmoid function 
to  normalize all weights between 0 and 1. Relying on this attention operator, two ways to plug it within an architecture are subsequently investigated.\\
    \textbf{(Single)}  \textbf{single branch:}  the classical approach which feeds the final classification layers with the weighted features $ \alpha_i  h_i$ preserving their dimension.\\
    \textbf{(Multi)}  \textbf{multi branch:} a multi-head model approach with an auxiliary head trained simultaneously on the same task. Its classifier inputs are average features $\bar{h} = \sum_{i=1}^{B}{\alpha_i h_i}$, reducing therefore dimensions and computation costs. It provides a view of the network decision based on the extracted features.

Regarding the proposed models, they are denoted using \textbf{Sdeep} as a prefix and listed in Table~\ref{tab:perf_model}. \textbf{Sdeep} stands for \textit{START Deep}, the CNES project that funded this work.

\section{Experiments} \label{sec:RESULTS}

\subsection{Dataset and preprocessing}
The Sentinel-2 SITS covering the Réunion island, already used to assess the \textit{TASSEL} model \cite{IENCOTassel}, is employed in this paper for easier comparisons. The land cover ground truth has been made available in~\cite{DUPUY2020}. This SITS consists of 21 images acquired between January and December 2017 with a 10 m spatial resolution. Four spectral bands are considered: B2 (blue), B3 (green), B4 (red) and B8 (near-infrared). In addition, two standard LCC indexes serve as synthetic channels, namely the Normalized Difference Vegetation Index (NDVI) 
and the Normalized  Difference Water Index (NDWI) \cite{McFeeters1996}. They are expressed as $NDVI=f(B8,B4)$ and $NDWI=f(B3,B8)$, where $f$ is the homogeneous function from $\mathbb{R_+^*}\times \mathbb{R_+^*}$ to $[-1;1]$ such that $f(x,y) = \frac{x - y} {x + y}$.
These indexes are first computed before being rescaled along with others channels between 0 and 1.\\
The observed area corresponds to a 59 km  x 67 km scene described by 39M pixels. Among these pixels, 2\% (880.000 pixels) are annotated according to 11 land cover classes. As shown in  Table~\ref{tab:precision-recall}, classes are unbalanced.
Clouds are filtered using a mutlilinear interpolation \cite{IENCOTassel}.
As usually observed for Sentinel-2 data
, B2, B3 and B4 are highly correlated (c\textgreater 0.92). Channels B8, NDVI and NWI are also correlated (c\textgreater0.64). 
Therefore, one may consider two groups of correlated channels and design classification models benefiting from these specific behaviors.\\
\subsection{Experimental settings}
Annotated pixels are shuffled and then split into a training dataset (60\%), a validation and test dataset (20\% each), preserving similar class ratios. 
Pixels belonging to a same object all belong to the same dataset.
The performances of the proposed architectures are assessed against a classical random forest (500 trees, 200 splits), \textit{TASSEL} \cite{IENCOTassel} (10578 objects, 2 clusters per object), and \textit{TempCNN} \cite{Pelletier}.
For all neural networks,  the classical unweighted categorical 
cross-entropy $CE$ is considered, and the multi-head
loss is expressed as $L_{global}=CE(Y,Y_{main}) + \lambda CE(Y, Y_{aux})$
where $Y_{main} $ and $Y_{aux}$ are the model main and auxiliary outputs and $\lambda$ is an hyper-parameter controlling the importance of the auxiliary classification in the learning process. If no auxiliary output is present, then $\lambda=0$, $\lambda =0.5$ otherwise. This loss is monitored to select the best network configurations.  All gradients are back-propagated through an Adamgrad optimizer and an $\mathbb{L}^2$-regularization with a weight decay of $1.10^{-6}$  to avoid overfitting on all layers.\\
\subsection{Quantitative and qualitative results} 

\begin{figure}[htb]
    \includegraphics[width=\linewidth]{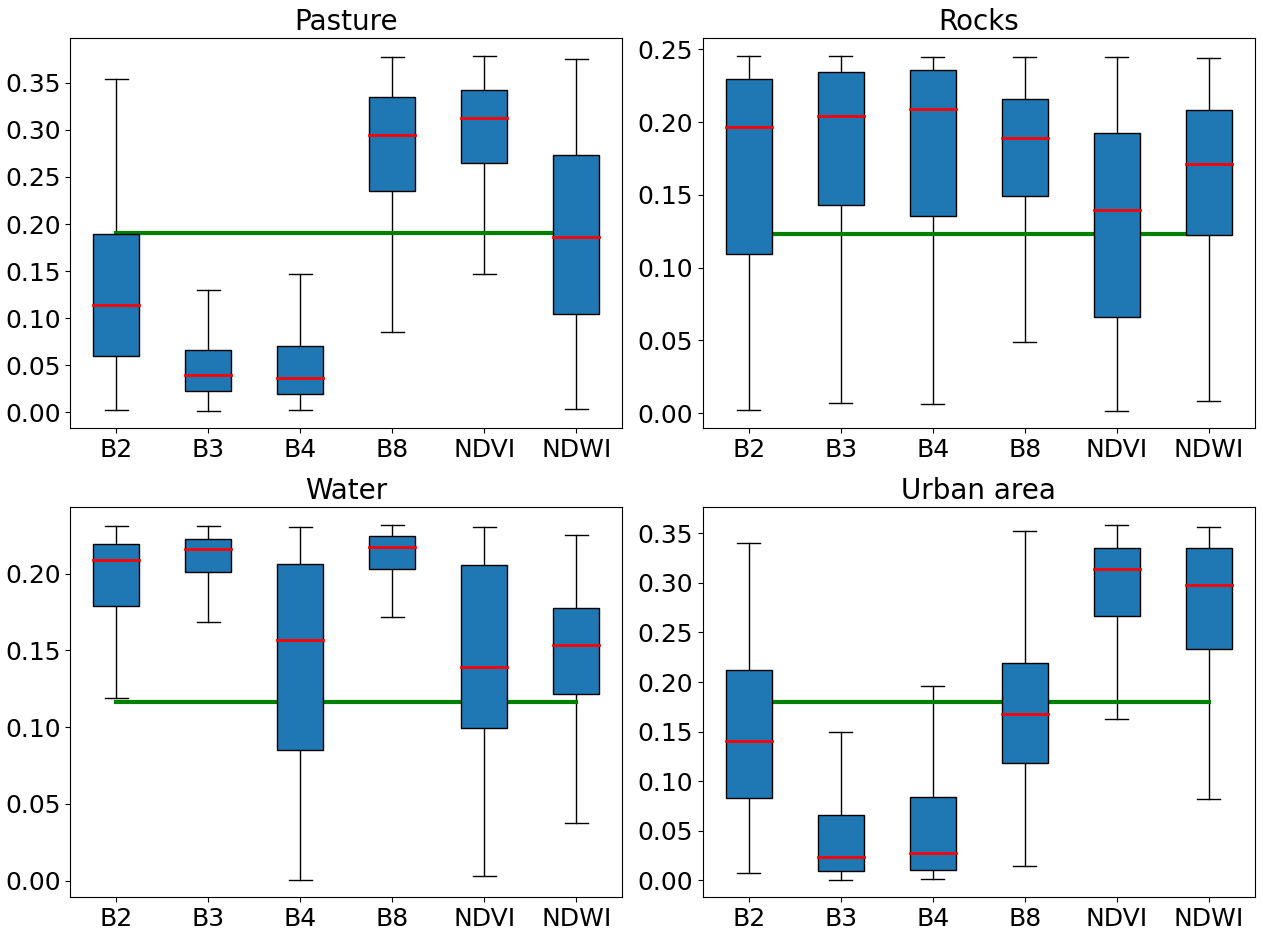}
    \caption{Boxplots of channel weight attention for 4 classes, normalized by class sums. Green horizontal lines depict activation thresholds (0.5 for a sigmoid).} 
    
    \label{fig:AttentionByClass}
\end{figure}

As observed in Table~\ref{tab:perf_model}, the reference models (Random Forest, \textit{TASSEL}, \textit{TempCNN}) all achieve good performances. Regarding the proposed architectures, similar or better performances are reached. The multi-head attention of \textbf{Sdeep-A-Multi-i} (91.1\%) provides slightly better results than those obtained with the single attention of \textbf{Sdeep-A-Single-i} (89.7\%). 
The classical features weighting approach adopted for the single branch architecture 
seems to degrade performance.
The two best performances, \textbf{Sdeep-B-Multi-ii} (92.2\%)  and \textbf{Sdeep-C-Multi-ii} (92.3\%), are obtained when processing the SITS at the original temporal scale, without any sub-sampling with the (ii) strategy, and by taking into account input channel correlations as for (B) and (C) features extraction types. Model \textbf{Sdeep-B-Multi-ii} is retained since it almost has twice less parameters for the same performance level.
Its performances are detailed in Table~\ref{tab:precision-recall}.
Class Relief Shadow and class Water are well detected, which is not the case for class Greenhouse crops. It can be mainly explained by the strong class imbalance 
(only $1,931$ pixels). 
Fig.~\ref{fig:AttentionByClass} shows the attention weights of the channel for each class. Beside allowing to discriminate classes, they bring meaningful information expressing to which extent
each channel features contribute to the decision, whatever its values, when used in such an architecture.
For instance B4 (red) has less importance on average, except for Water and Rocks. The NDVI and NDWI channels have more impact. For example, in class Water, the NDWI is obviously mobilized, but NDVI also contributes to the decision: its values (not its weight attention) are negative on average for that class, which can be indeed associated with the presence of water.
Regarding class Pasture, NDVI is obviously taken into account along with B8 as it exhibits vegetation. Finally, as expected for this zone, class Urban area fairly benefits from all bands.

\begin{table}[htb]
\centering
\resizebox{\linewidth}{!}{%
\begin{tabular}{|l|l|l|r|c|}
\hline
\textbf{Model} & 
\textbf{Architecture} &
\textbf{Conv. blocks} &
\multicolumn{1}{l|}{\textbf{\begin{tabular}[c]{@{}l@{}}Number of\\ parameters\end{tabular}}} &
\multicolumn{1}{l|}{\textbf{\begin{tabular}[c]{@{}l@{}}Test\\  accuracy\end{tabular}}} \\ 
\hline

Random Forest & \hfill -\hfill\null &\hfill -\hfill\null &\hfill -\hfill\null & 90.4 \\ \hline 
TASSEL & (A) + Multi &  (i), 6x(3)  & 3,647,510   & 89.5 \\ \hline 
TempCNN & (A) & (ii), 3x(9,1) & 807,284   & 91.3 \\ \hline
Sdeep-A-Multi-i & (A) + Multi & (i), 2x(7) & 14,346,262  & 91.1 \\ \hline 
Sdeep-A-Single-i & (A) + Single & (i), 2x(7) & 14,340,619  & 89.7 \\ \hline 
\color{green}Sdeep-B-Multi-ii & \color{green}(B) + Multi & {\color{green}(ii), 3x(9,1)} & \color{green}1,376,203 & \color{green}92.2 \\ \hline 
Sdeep-C-Multi-ii  & (C) + Multi & (ii), 3x(9,1) & 2,445,256  & 92.3 \\ \hline 

\end{tabular}%
}
\caption{Accuracy for the reference and proposed Sdeep models. Architecture: features extraction type (A), (B) or (C) + Single or Multi branch attention.  Conv. blocks: type (i) or (ii), number of chained convolutions x (kernel shape).}
\label{tab:perf_model}
\end{table}

\begin{table}[ht]
    \centering
    \resizebox{0.9\linewidth}{!}{%
    \begin{tabular}{|l|l|l|c|}
    \hline
    \textbf{Class} & \textbf{Precision} & \textbf{Recall} & \textbf{\% of annotated pixels} \\ \hline
    Sugar cane & 96.7 & 96.6 & 12.4 \\ \hline
    Pasture & 92.8 & 94.0 & 7.3 \\ \hline
    Market gardening & 75.1 & 74.3 & 2.3 \\ \hline
    \color{red}Greenhouse crops & \color{red} $\underline{52.9}$& $\color{red}\underline{52.3}$ & \color{red}0.2 \\ \hline
    Orchards & 80.1 & 83.7 & 3.9 \\ \hline
    Wooded areas & 87.3 & 94.4 & 23.5 \\ \hline
    Moor & 92.4 & 77.9 & 16.0 \\ \hline
    \color{green} Rocks & \color{green}$\overline{97.5}$ & $\color{green}\overline{97.7}$ &  \color{green}21.4 \\ \hline
    Relief shadows & 94.3  & 98.7 & 5.1 \\ \hline
    \color{green} Water & $\color{green}\overline{99.9}$ & $\color{green}\overline{99.4}$ & \color{green} 6.1 \\ \hline
    Urban area & 84.6 & 91.0 & 1.8 \\ \hline
    \textbf{Mean} & \textbf{86.7} & \textbf{87.3} & \hfill - \hfill \\ \hline
    \end{tabular}%
    }
    \caption{Precision and recall by class for model \textbf{Sdeep-B-Multi-ii} on the test set (170,000 pixels). Last column shows the ratios of annotated pixels.} 
    \label{tab:precision-recall}
\end{table}
\vspace{-2mm}


\section{CONCLUSION}

In this paper, an attention-based DNN is proposed to 
perform a LCC task using a SITS.
Predictions are explained by determining the contribution of each input channel to the final decision. This DNN
relies on an auxiliary attention mechanism and temporal convolutional layers. The latter take into account channel correlations 
to lower the number of parameters.
Obtained results show that meaningful interpretations of the outcomes are provided while reaching state-of-the-art classification performances. Future works include conducting a more extensive qualitative evaluation and assessing the relevance of the proposed architecture on other SITS, whether Sentinel-2 ones or not.

\bibliographystyle{IEEE}
\bibliography{biblio}

\end{document}